\documentclass{article}
\usepackage{xcolor}
\usepackage{spconf,amsmath,graphicx}
\usepackage[english]{babel}
\usepackage[utf8]{inputenc}
\usepackage{algorithm}
\usepackage[noend]{algpseudocode}
\usepackage{booktabs}
\usepackage{footnote}
\usepackage{multirow}
\makesavenoteenv{tabular}
\makesavenoteenv{table}


\title{Robust MAML: Prioritization task buffer with adaptive learning process for model-agnostic meta-learning}

\name{Thanh Nguyen, Tung Luu, Trung Pham, Sanzhar Rakhimkul, Chang D. Yoo}
\address{Korea Advanced Institute of Science and Technology (KAIST)}

\begin{document}
%
\maketitle
\begin{abstract}
Model agnostic meta-learning (MAML) is a popular state-of-the-art meta-learning algorithm that provides good weight initialization of a model given a variety of learning tasks. The model initialized by provided weight can be fine-tuned to an unseen task despite only using a small amount of samples and within a few adaptation steps. MAML is simple and versatile but requires costly learning rate tuning and careful design of the task distribution which affects its scalability and generalization. This paper proposes a more robust MAML based on an adaptive learning scheme and a prioritization task buffer (PTB) referred to as Robust MAML (RMAML) for improving scalability of training process and alleviating the problem of distribution mismatch. RMAML uses gradient-based hyper-parameter optimization to automatically find the optimal learning rate and uses the PTB to gradually adjust training task distribution toward testing task distribution over the course of training. Experimental results on meta reinforcement learning environments demonstrate a substantial performance gain as well as being less sensitive to hyper-parameter choice and robust to distribution mismatch\let\thefootnote\relax\footnotetext{This work was supported by Institute for Information \& communications Technology Planning \& Evaluation (IITP) grant funded by the Korea government (MSIT) (Projects: No. 2017-0-01780 and No. 2019-0-01396)}.
\end{abstract}

\section{Introduction}

Meta-learning, often referred to as learning to learn, has emerged as a potential learning paradigm that can absorb information from tasks and generalize that information to unseen tasks proficiently. It makes learning more general: efforts being made to construct task distributions, from which meta-learning algorithms can automatically provide good weight initialization and a set of hyper-parameters. A relatively recent landmark  meta-learning algorithm is Model-Agnostic Meta-Learning (MAML) \cite{finn2017model} which is a conceptually simple, general algorithm that has shown impressive results over many problems ranging from few-shot learning problems in classification, regression, and reinforcement learning (RL) \cite{antoniou2018train}. MAML trains a model on a task distribution to acquire the optimal weight initialization. The model then can be adapted to an unseen task with few sample and few adaptation steps - often in one step \cite{finn2017model}.

\begin{figure}[t]
    \begin{minipage}[b]{1\linewidth}
      \centering
      \centerline{\includegraphics[width=9.0cm]{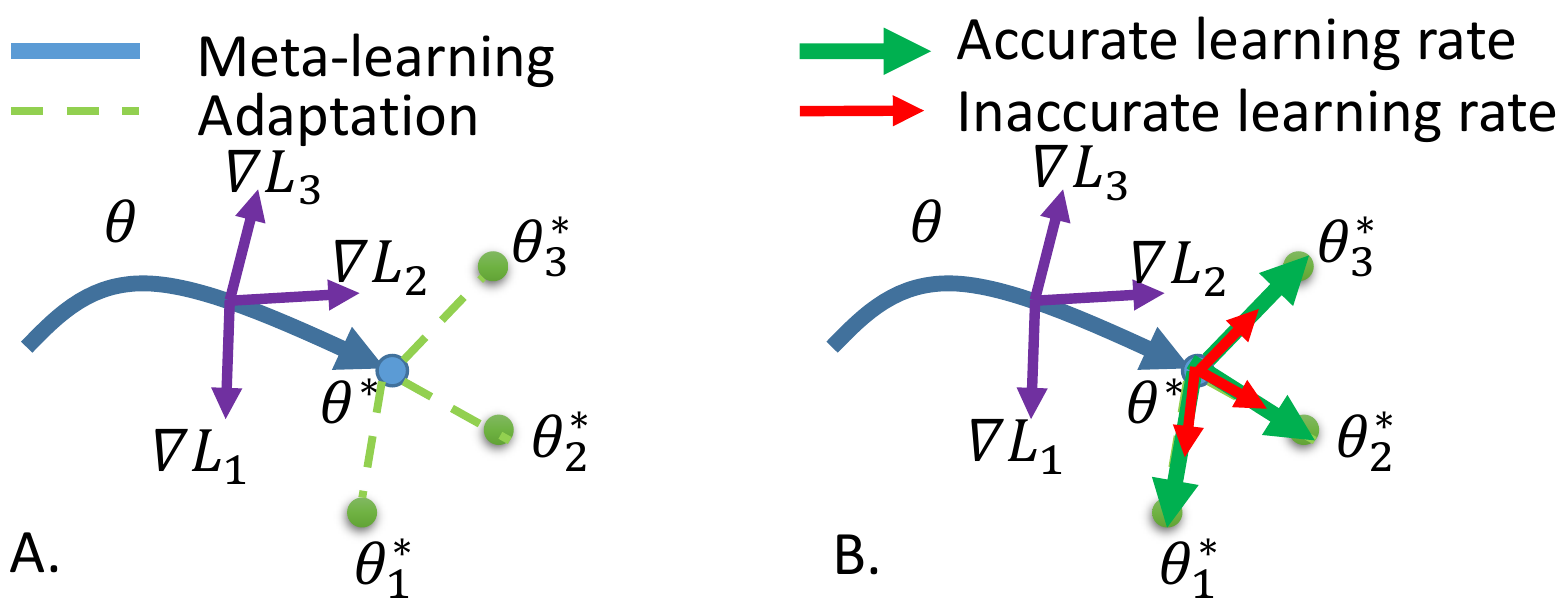}}
    \end{minipage}
    \caption{\textbf{A}. shows how MAML learns weight initialization \cite{finn2017model}. \textbf{B}. shows the effect on the adaptation process of different learner learning rates.}
    \label{fig:maml_1}
\end{figure}

Although MAML has an interesting approach, the generality and simplicity of the algorithm come with two difficulties. First, MAML is considered expensive in terms of computational cost since the model is trained on multiple tasks and requires the computation of the second-order derivative. Moreover, after training, the model is adapted to an unseen task with a few samples and a few adaptation steps that require an accurate learner learning rate. Thus, the time for searching hyper-parameters significantly increases compared to other meta-learning approaches. The difference between the accurate and the inaccurate learning rate of the learner is shown in Fig \ref{fig:maml_1}, intuitively. The sensitivity of hyper-parameters to performance can lead to poor scalability. Second, learning the weight initialization requires careful design of task distribution to achieve high performance. Ideally, task representations in latent space should collapse into one mode and the unseen task should conform well to the task distribution. In reality, even with good prior knowledge of the tasks, it is difficult to manually design the perfect task distribution. A slight mismatch in distribution between training tasks and testing tasks distribution can lead to poor generalization in the adaptation process \cite{mehta2020curriculum}. Thus, uniformly sampling from the training task distribution, in the manner done by MAML, is generally considered to be an inefficient strategy.

This paper proposes Robust MAML (RMAML): an adaptive learning scheme and a prioritization task buffer (PTB) to address the two aforementioned problems. For hyper-parameter tuning, the learner learning rate can be gradually adjusted to minimize the validation loss over the course of training automatically instead of manual tuning. The learner learning rate can vary among weights or layers of the model to provide more flexibility to the adaptation process without any human effort thanks to the adaptive learning scheme. For distribution mismatch, besides uniformly sampling tasks from task distribution, PTB interferes with the training process by providing additional specified tasks prioritized on validation return. This helps to correct the training task distribution to align with testing task distribution assuming the training task distribution is the unimodal distribution with noise and the testing task distribution aligns with the free noise training task distribution. The validation return is available in meta-learning algorithms that require no extra effort. RMAML shows good performance, helps to stabilize the training process, increases the scalability, and robustness to distribution mismatch.

\section{Related work}
 Meta-learning involves learning a learning algorithm that can adapt quickly to an unseen task with few samples. Meta learning papers can be classified into the following three approaches: metric-based  \cite{koch2015siamese}\cite{vinyals2016matching}\cite{sung2018learning}\cite{snell2017prototypical},  model-based \cite{santoro2016meta}\cite{graves2014neural}\cite{munkhdalai2017meta}, and  optimization-based \cite{nichol2018first}\cite{finn2018probabilistic}. This paper directly falls into the category of the optimization-based approach which focuses on the optimization process by customizing the optimizer \cite{ravi2016optimization} or finding good weight initialization (MAML \cite{finn2017model}). MAML has shown impressive performance but also has some problems. Previously, efforts have been made to reduce computation cost of MAML by using the first-order approximation of gradient \cite{nichol2018first}, raising problems about task distribution sensitivity and using reinforcement learning to correctly choose training tasks \cite{mehta2020curriculum}, or providing adaptation model in a probabilistic way \cite{finn2018probabilistic}. By contrast, RMAML solves the scalability caused by costly hyper-parameter tuning and task distribution mismatch between training and testing. 

The issues discussed in this paper are closely related to the topic hyper-parameter optimization which focuses on finding the best set of hyper-parameters automatically. Many methods have been proposed ranging from naive grid search to more advanced approaches such as using Bayesian \cite{snoek2012practical}, model-based \cite{bergstra2013making}, reversible learning \cite{maclaurin2015gradient}, or hypergradient descent \cite{baydin2017online}. Leveraging the success of this field but keeping the method lightweight, simple and effective, RMAML chooses gradient-based optimization for hyper-parameter by minimizing the validation loss which is conveniently available in the training process of MAML.
\begin{algorithm}
    \small
    \caption{MAML for Reinforcement Learning}
    \label{alg:maml}
    \begin{algorithmic}[1]
    \Require $p(\mathcal{T}):$ distribution over tasks
    \Require $\alpha, \beta:$ step size hyper-parameters
    \State Randomly initialize $\theta$
    \State Initialize $\alpha = \alpha_0$
    \While{not done}
    \State \text {Uniformly sample M tasks } $\mathcal{T}_{i} \sim p(\mathcal{T})$
    \ForAll{$\mathcal{T}_{i}$} 
    \State Sample K trajectories $D^i_{train}$ using $f_\theta$ in $\mathcal{T}_{i}$
    \State Update one-step gradient descent using $D^i_{train}$
    \State $\theta_{i}^{\prime}=\theta-\alpha \nabla_{\theta}  \mathcal{L}^{train}_{\mathcal{T}_{i}}\left(f_{\theta}\right)$ 
    \State Sample trajectories $D^i_{val}$ using $f_{\theta'_i}$ in $\mathcal{T}_{i}$
    \EndFor
    \State Update initialization weight using each $\mathcal{D}^{i}_{val}$
    \State $\theta \leftarrow \theta-\beta \nabla_{\theta} \sum_{\mathcal{D}_{val}} \mathcal{L}^{val}_{\mathcal{T}_{i}}\left(f_{\theta_{i}^{\prime}}\right)$ 
    \EndWhile
    \end{algorithmic}
\end{algorithm}

RMAML is performed on Reinforcement Learning (RL) tasks since it is considered the most challenging problem for meta-learning, also known as meta reinforcement learning. The goal is to find a policy that can quickly adapt to an unseen environment from only a few trajectories. Some methods try to solve it by conditioning the policy on a latent representation of the task \cite{rakelly2019efficient} or using a recurrent neural network \cite{duan2016rl}. MAML has also shown some achievements utilizing REINFORCE loss \cite{williams1992simple} for inner adaptation loop and TRPO \cite{schulman2015trust} for outer meta-learning loop.

\section{MAML}

 MAML is a optimization-based meta learning algorithm that learns an optimal weight initialization for a model given a task distribution. Formally, the model $f_\theta$ is parameterized by $\theta$. Given a task generated from task distribution $ P $  ${\tau_i} \sim P(\tau)$ and its associated training and validation datasets are $\left(\mathcal{D}_{train }^{i}, \mathcal{D}_{val}^{i}\right)$. The model can be trained by one or more gradient descent steps (adaptation steps) as follows
\begin{equation*}
    \theta_i^{\prime}=\theta-\alpha \nabla_{\theta} \mathcal{L}_{\tau_{i}}^{train}\left(f_{\theta}\right),
\end{equation*}
where $\mathcal{L}_{\tau_{i}}^{train}$ is the loss for task $\tau_i$ computed using $D_{train}^i$. Here $\alpha$ is the learning rate of the learner. To achieve stable generalization across $P$, MAML finds the optimal initialization weight $\theta^*$ such that the task-specific fine-tuning achieves low validation loss. The solution can be acquired by minimizing the validation after adapting across $\tau_i$ computed using $\mathcal{D}_{\text {val}}^{i}$:
\begin{align*}
    \theta^{*} &=\arg \min_{\theta} \sum_{\tau_i \sim p(\tau)} \mathcal{L}_{\tau_i}^{val} \left(f_{\theta^{'}_i}\right)\\
     &=\arg \min _{\theta} \sum_{\tau_{i} \sim p(\tau)} \mathcal{L}_{\tau_{i}}^{val}\left(f_{\theta-\alpha \nabla_{\theta} \mathcal{L}_{\tau_{i}}^{train}\left(f_{\theta}\right)}\right),
\end{align*}
\begin{equation}
     \theta \leftarrow \theta-\beta \nabla_{\theta} \sum_{\tau_{i} \sim p(\tau)}
     \mathcal{L}_{\tau_{i}}^{val}\left(f_{\theta-\alpha \nabla_{\theta} \mathcal{L}_{\tau_{i}}^{train}\left(f_{\theta}\right)}\right), \nonumber
     \end{equation}
where $\beta$ is the meta learning rate and $\mathcal{L}_{\tau_i}^{val}$ is the validation loss for $\tau_i$. For RL meta-learning, $\mathcal{L}^{train}$ is REINFORCE loss \cite{williams1992simple} and $\mathcal{L}_{\tau_i}^{val}$ is equivalent loss used in TRPO \cite{schulman2015trust}. Each meta update sample $M$ tasks, referred to as meta-batchsize. The pseudo-code of MAML for RL is shown in Algorithm \ref{alg:maml}.

\section{Robust MAML}
\label{sec:pagestyle}
 Instead of using a fixed learner learning rate as hyper-parameter, RMAML optimizes the learner learning rate by gradient descent to minimize the evaluation loss. Optionally, the learning rate can vary among weights or layers to provide more flexibility in the adaptation process. In RMAML, one-step gradient of the learner learning rate is given as:\\
\begin{align*}
\frac{\partial \mathcal{L}^{val}_{\mathcal{T}_{i}}\left(f_{\theta_{i}^{\prime}}\right)}{\partial \alpha} &= \frac{\partial \mathcal{L}^{val}_{\mathcal{T}_{i}}\left(f_{\theta_{i}^{\prime}}\right)^{T}}{\partial \theta_{i}^{\prime}} \frac{\partial \theta_{i}^{\prime}}{\partial \alpha} \\
&=\nabla_{\theta_{i}^{\prime}} \mathcal{L}^{val}_{\mathcal{T}_{i}}\left(f_{\theta_{i}^{\prime}}\right)^{T}\frac{\partial\left(\theta-\alpha \nabla_{\theta} \mathcal{L}_{\mathcal{T}_i}^{train}\left(f_{\theta}\right)\right)}{\partial \alpha} \\
&=\nabla_{\theta_{i}^{\prime}} \mathcal{L}_{\mathcal{T}_i}^{val}\left(f_{\theta_{i}^{\prime}}\right)^T \cdot\left(-\nabla_{\theta} \mathcal{L}_{\mathcal{T}_ i}^{train}\left(f_{\theta}\right)\right).
\end{align*}
\begin{equation}
\alpha=\alpha + \alpha_{0} \sum_i \nabla_{\theta_{i}^{\prime}} \mathcal{L}_{\mathcal{T}_i}^{val}\left(f_{\theta_{i}^{\prime}}\right)^T \nabla_{\theta} \mathcal{L}_{\mathcal{T}_ i}^{train}\left(f_{\theta}\right) \nonumber
\end{equation}
Notice that, although  $\alpha$ is substituted by another parameter  $\alpha_0$, $\alpha_0$ is not sensitive to the performance of the algorithm and can be safely set at small value (e.g 1e-2). The meta learning rates can also be adjusted. However, the meta learning rate is not sensitive to the performance compared to the learner learning rate. Therefore, the meta-learning rate is set equal to $\alpha_0$.

In terms of training task distribution $P$, MAML uniformly samples task from $P$ (UNIFORM STRATEGY). It performs well with uni-modal distribution but is sensitive to noisy tasks. Intuitively, the noisy tasks, far from the mode of $P$, can pull the solution away from the optimal initialization weight. Actively training models more on useful tasks yields better convergence. The problem is that $P$ can not be accessed directly. Querying task information consumes computation cost and the information highly depends on the querying RL policy. Wisely using the task information is crucial. RMAML introduces  a prioritization task buffer (PTB), denoted by $B$.  PTB adds more useful tasks during training besides uniformly sampling from the task distribution $P$ as follows: at every meta update, all $M$ training tasks are kept in $B$ with corresponding validation returns. In the next iteration, $M$ tasks, required for learner update,  are $L$ tasks sampled from $B$ plus $(M-L)$ tasks uniformly sampling from $P$. After sampling, $B$ is cleared to prepare it to receive new tasks. The $L$ tasks, whose validation return is relatively medium in $B$, are chosen. $L$ is gradually increased from 0 to $MAX_L$ ($\leq M$) over the course of training (MEDIUM STRATEGY).

\begin{figure}[htb]
    \begin{minipage}[b]{1\linewidth}
      \centering
      \centerline{\includegraphics[width=8.8cm]{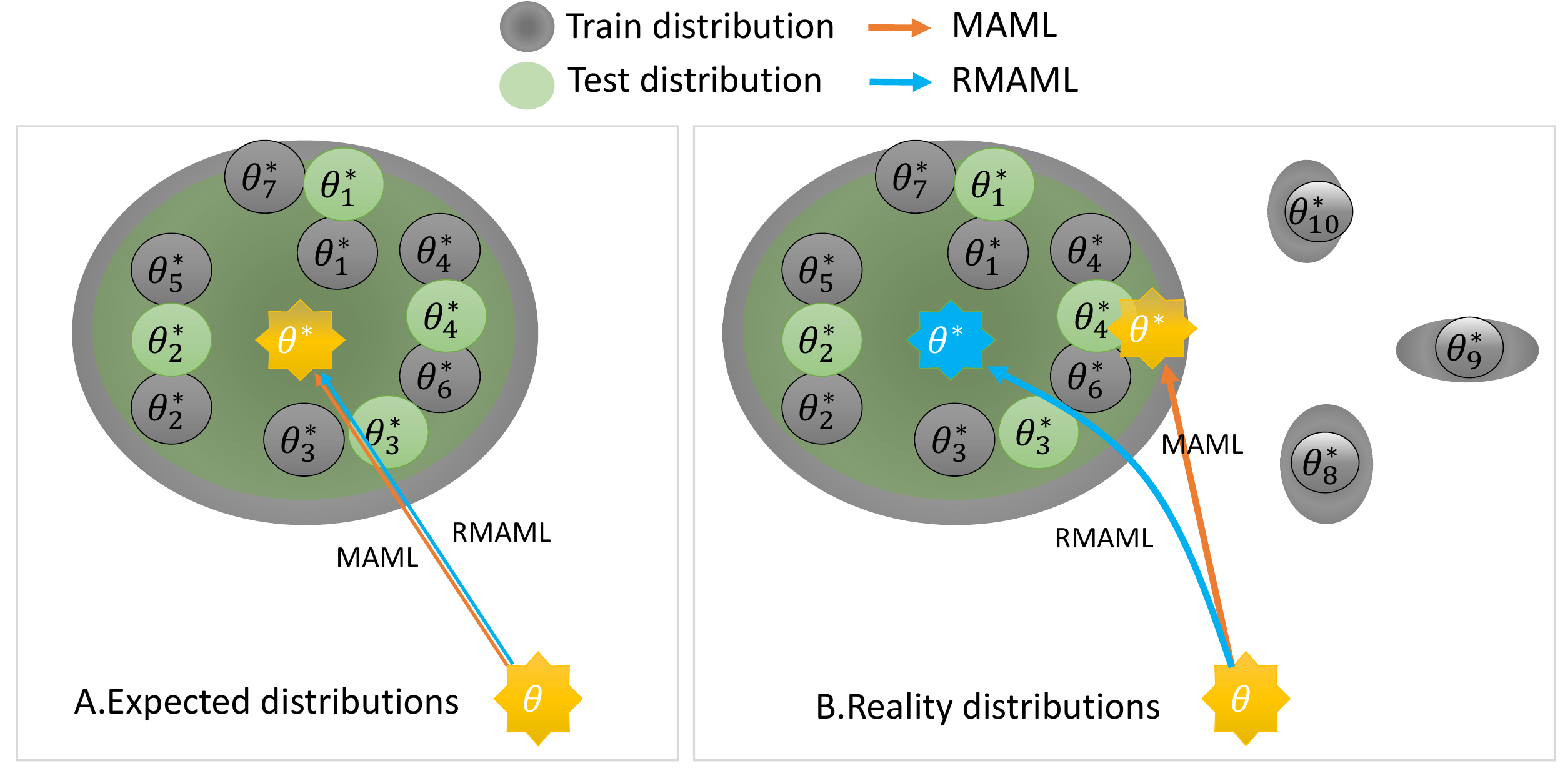}}
    \end{minipage}
    \caption{The demonstration of distribution mismatch between train/test and the corresponding behavior of MAML/RMAML.}
    \label{fig:rmaml}
\end{figure}

Fig \ref{fig:rmaml} shows the behaviour of MAML/RMAML in two type distributions. Ignoring the bias inducing by noisy tasks, the majority of useful tasks will gradually pull the resulting weight towards the optimal weight. PTB helps to increase the rate of training on the useful tasks to eliminate bias. Furthermore, the better the RL policy the more trust-able the buffer since validation return is more distinguishable between useful tasks and noisy tasks. Using a small number of $L$ in the beginning of the training and gradually increase its value showed better performance. Regarding how to choose useful tasks in $B$, an interesting empirical observation is that the tasks with the lower validation loss (EASY STRATEGY) tend to move the policy to a local minimums, whereas tasks with higher validation loss (HARD STRATEGY) will move policy somewhere far from the optimum. Tasks with medium validation loss (MEDIUM STRATEGY) successfully move the policy closer to the optimum. This strategy is consistent with curriculum learning which proposes training on tasks that are not too easy and also not too hard \cite{bengio2009curriculum}. The pseudo-code is shown in Algorithm \ref{alg:rmaml}.

\begin{algorithm}
\small
\caption{RMAML for Reinforcement Learning}
\label{alg:rmaml}
\begin{algorithmic}[1]
\Require $p(\mathcal{T}):$ distribution over tasks
\Require  $\alpha_0$ step size hyper-parameter
\Require Initialize Priority Task Buffer B 
\State Randomly initialize $\theta$
\While{not done} 
\State Sample L tasks  $\mathcal{T}_B \sim B$ \Comment{MEDIUM STRATEGY}  
\State Uniformly Sample $(M-L)$ tasks  $\mathcal{T}_P \sim p(\mathcal{T})$
\State $\mathcal{T}_{i} =  \mathcal{T}_B \text{  concat  } \mathcal{T}_P $
\State Empty B 
\ForAll{$\mathcal{T}_{i}$} 
\State Sample K trajectories $D^i_{train}$ using $f_\theta$ in $\mathcal{T}_{i}$
\State Update one-step gradient descent using $D^i_{train}$
\State $\theta_{i}^{\prime}=\theta-\alpha \nabla_{\theta}  \mathcal{L}^{train}_{\mathcal{T}_{i}}\left(f_{\theta}\right)$ 
\State Sample trajectories $D^i_{val}$ using $f_{\theta'_i}$ in $\mathcal{T}_{i}$
\EndFor
\State Update initialization weight using each $\mathcal{D}^{i}_{val}$
\State $\theta \leftarrow \theta-\alpha_0 \nabla_{\theta} \sum_{\mathcal{D}_{val}} \mathcal{L}^{val}_{\mathcal{T}_{i}}\left(f_{\theta_{i}^{\prime}}\right)$ 
\State $\alpha=\alpha+\sum_i \alpha_{0} \nabla_{\theta_{i}^{\prime}} \mathcal{L}_{\mathcal{T}_i}^{val}\left(f_{\theta_{i}^{\prime}}\right)^T \cdot\nabla_{\theta} \mathcal{L}_{\mathcal{T}_ i}^{train}\left(f_{\theta}\right)$
\State  Store all $\mathcal{T}_{i} ,\mathcal{L}^{val}_{\mathcal{T}_{i}}\left(f_{\theta_{i}^{\prime}}\right) \rightarrow  B$
\EndWhile
\end{algorithmic}
\end{algorithm}

\section{Experiments}
\label{sec:typestyle}
\textbf{Sampling Strategy Investigation}. Our team designed a environment called 2D REACHING which consists of 300 tasks that are drawn from  a mixture of three independent normal distributions. Each task corresponds to a 2D point which a RL policy is required to reach. The largest normal distribution contains 200 points (assuming it matches the testing distribution). Each remaining normal distribution contains 50 tasks playing the role as noise tasks. Assume we have a perfect MAML algorithm that can move the agent toward the given task proportional to the distance between them. The weight of the agent is the current position of the agent. Prioritization scores are the distance between tasks and the updated agent position. Different PBT strategy are investigated: EASY, HARD, MEDIUM, UNIFORM.  The goal is to make MAML produce the weight near the center's largest normal distribution. 

\begin{figure}[htb]
\begin{minipage}[b]{1.0\linewidth}
  \centering
  \centerline{\includegraphics[width=9.0cm]{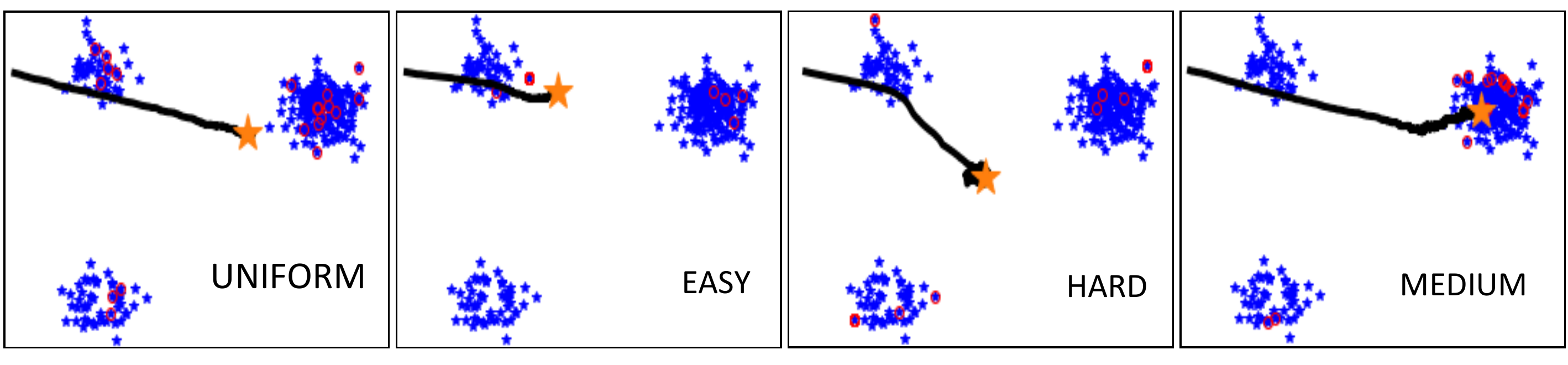}}
\end{minipage}
\caption{Different sampling strategies performed on the 2D REACHING environment. UNIFORM: random sampling EASY: use  PTB prioritizing easy tasks, HARD: use PTB prioritizing hard tasks, MEDIUM: use PTB prioritizing medium tasks.}\label{fig:sampling strategy}
\end{figure}

The results in Fig \ref{fig:sampling strategy} show that the MEDIUM strategy allows the robot to move to the desired destination without being stuck in local minima.

\noindent
\textbf{Reinforcement Learning Tasks}. To demonstrate the effectiveness, RMAML is evaluated on high dimensional locomotion tasks simulated in MuJoCo   \cite{todorov2012mujoco}. For fare comparison, the same setup in MAML \cite{finn2017model} is applied.  There are two tasks: Cheetah Velocity (VEL), Cheetah  Direction (DIR). In VEL, a Cheetah robot must run at a particular velocity, chosen uniformly at a random value between 0.0 and 2.0. In DIR, the Cheetah robot must run in a particular, randomly chosen direction (forward/backward). The hyper-parameters are used in the same way mentioned in MAML \cite{finn2017model}.

For RMAML specific hyper-parameter, we set $\alpha_0 =0.01$, L is gradually increased up to 1/4 of the meta-batch size. For implementation, to reduce the wall time, RMANL is implemented using distributed training which is currently popular in the RL field \cite{barth2018distributed}. For testing, 40 tasks are sampled randomly. The model is initialized by RMAML and evaluated over 40 tasks (roll out 20 trajectories for each task). The average return among trajectories and tasks is reported as step 0. Then, the model performs one step adaptation with gradient descent, rollout and the average return is reported as step 1. Table \ref{tab:res1} shows the average test return from different algorithms: RMAM, MAML+ (our MAML implementation with distributed training), MAML\cite{finn2017model} and Pretrain\cite{finn2017model}. The result shows that RMAML consistently outperforms MAML+ in both environments and outperform the original MAML in the VEL.
\begin{table}[t]
\centering
\small
\setlength{\tabcolsep}{4.5pt} 
\begin{tabular}{cc|cccc}
\toprule
Task                 & Step & Pretrain\cite{finn2017model} & MAML\cite{finn2017model}   & MAML+ \footnote{MAML reimplementation using distributed training} & RMAML \\ \midrule
\multirow{2}{*}{VEL} & 0    & -158.0   & -125.0 & -60.1 & \textbf{-58.0} \\
                     & 1    & -137.0   & -79.0  & -41.5 & \textbf{-31.2} \\ \midrule
\multirow{2}{*}{DIR} & 0    & -40.5    & -50.7  & 30.3  & \textbf{18.2}  \\
                     & 1    & -38.3    & \textbf{293.2}  & 215.7 & 272.9 \\ \bottomrule
\end{tabular}
\caption{Result of the RL locomotion tasks featuring the average test return on Half Cheetah Velocity {[}VEL{]}, Half Cheetah Direction {[}DIR{]}, 2D Navigation {[}2D{]} with 0 and 1 adaptation step.}
\label{tab:res1}
\end{table}
To verify the robustness of RMAML with distribution mismatch, the Cheetah Velocity is customized to be Noise Cheetah Velocity (NoiseVEL). NoiseVEL adds 20\%  noise tasks chosen uniformly at random between 3.0 and 4.0 during training. During test time we stop adding noise tasks and evaluate the algorithm as mentioned above. 

\begin{table}[htp]
\centering
\footnotesize
\begin{tabular}{cc|cc}
\toprule
Phase                          & Step & MAML+\footnotemark[\value{footnote}] & RMAML \\ \midrule
\multirow{2}{*}{Train (Noise)} & 0    & -99.1 & \textbf{-84.3} \\
                               & 1    & -71.7 & \textbf{-60.4} \\ \midrule
\multirow{2}{*}{Test}          & 0    & -60.3 & \textbf{-42.0} \\
                               & 1    & -54.9 & \textbf{-31.8} \\ \bottomrule
\end{tabular}
\caption{Noise Cheetah Velocity Result featuring average test return during the training phase and testing phases with and without noise.}
\label{tab:res2}
\end{table}

The results in Table \ref{tab:res2} shows that RMAML outperforms MAML+ on both training and testing as well as reach to near -31.2 similar to training on VEL which has no noise.
\section{Conclusion}
This paper presents Robust MAML, a prioritization task buffer with an adaptive learning process for model-agnostic meta-learning. RMAML substantially reduces hyper-parameter tuning time and it is robust to distribution mismatch. This makes RMAML suitable and effective to scale for a variety of problems. RMAML shows consistent results that outperforms the original MAML locomotion meta-RL benchmarks.




\vfill\pagebreak
\bibliographystyle{ieeebib}
\bibliography{main}

\end{document}